\documentclass[10pt,twocolumn,letterpaper]{article}

\usepackage{iccv}
\usepackage{times}
\usepackage{epsfig}
\usepackage{graphicx}
\usepackage{amsmath}
\usepackage{amssymb}

\usepackage[breaklinks=true,bookmarks=false]{hyperref}

\iccvfinalcopy 

\def\iccvPaperID{****} 

\ificcvfinal\fi

\begin{document}

\title{Few-shot Knowledge Transfer for Fine-grained Cartoon Face Generation}

\author{
Nan Zhuang$^{1,2}$, Cheng Yang$^{2}$ \\
$^{1}$Peking University, Beijing, China \\
~$^{2}$ByteDance AI Lab, Beijing, China \\
{\tt\small zhuangn53@pku.edu.cn, yangcheng.iron@bytedance.com} \\
}


\maketitle
\ificcvfinal\thispagestyle{empty}\fi

\begin{abstract}
  In this paper, we are interested in generating fine-grained cartoon faces for various groups. We assume that one of these groups consists of sufficient training data while the others only contain few samples. Although the cartoon faces of these groups share similar style, the appearances in various groups could still have some specific characteristics, which makes them differ from each other. A major challenge of this task is how to transfer knowledge among groups and learn group-specific characteristics with only few samples. In order to solve this problem, we propose a two-stage training process. First, a basic translation model for the common group (which consists of sufficient data) is trained. Then, given new samples of other groups, we extend the basic model by creating group-specific branches for each new group. Group-specific branches are updated directly to capture specific appearances for each group while the remaining group-shared parameters are updated indirectly to maintain the distribution of intermediate feature space. In this manner, our approach is capable to generate high-quality cartoon faces for various groups.
\end{abstract}

\section{Introduction}

Cartoon faces can be seen everywhere in our daily life. They are widely used as profile pictures in social media platforms, such as QQ, WeChat, Weibo, Twitter, etc. Some interesting stickers or memes are also made up of cartoon faces. However, drawing a cartoon picture can be quite difficult. Given a real-face image, it may takes a professional painter several hours to finish such an animation artwork in order to keep personal characteristics. In this paper, we aim at generating a realistic cartoon face according to the given real-face image. We tackle this problem as an image-to-image translation task.

Image-to-image translation aims to learn a function that maps images within two different domains. This topic has gained a lot of attention from researchers in the field of machine and computer vision. It is first introduced by Isola \etal \cite{pix2pix}, which utilizes the generative adversarial network to learn the mapping function from paired training samples.  However, paired data is usually not easy to obtain. Thus CycleGAN \cite{cycleGAN} tries to learn translation models from unpaired data. They use the cycle consistency loss to learn two mapping functions at the same time and achieve impressive results.

\begin{figure}[t]
\begin{center}
\includegraphics[width=1.0\linewidth]{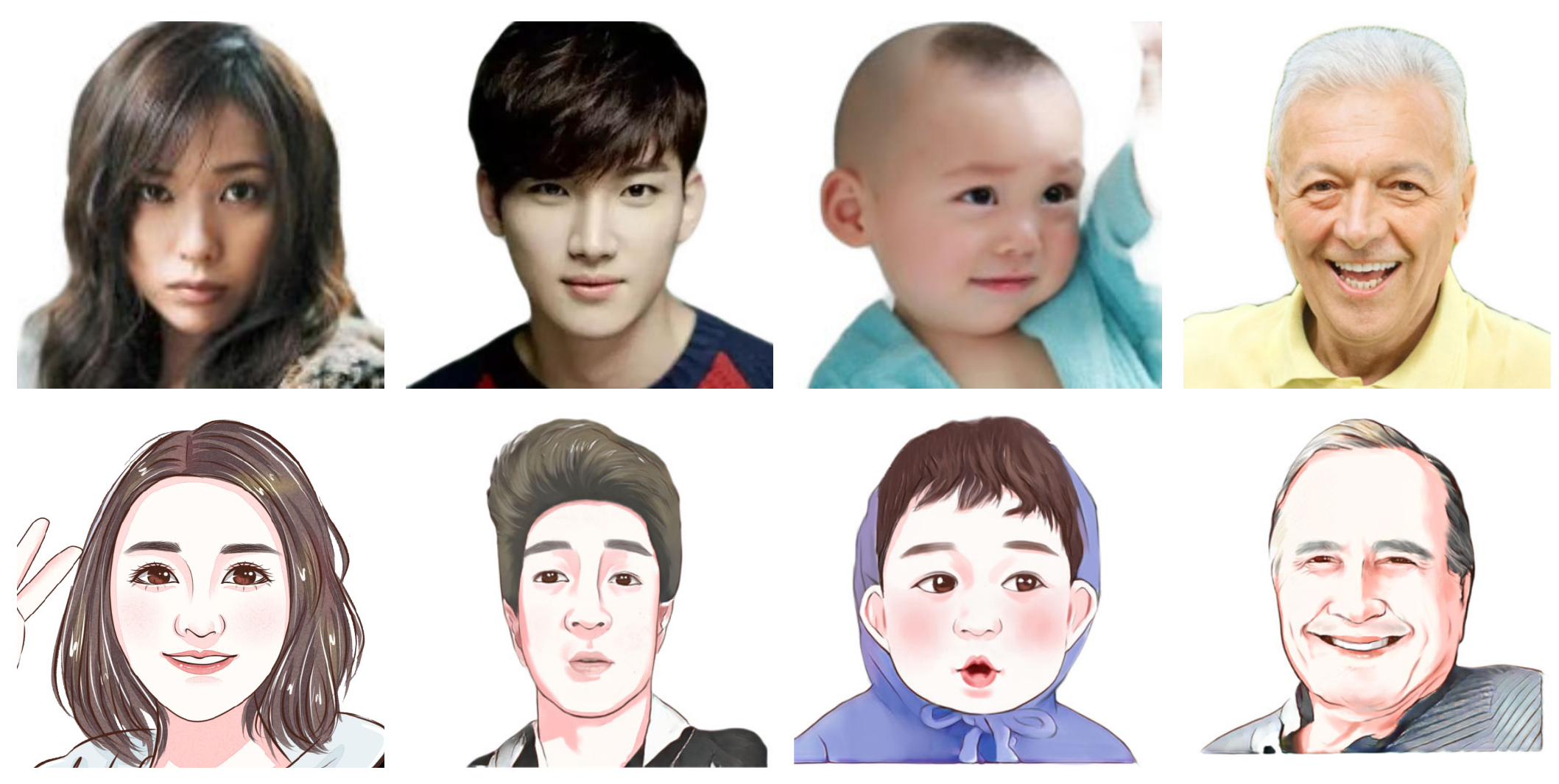}
\end{center}
\vspace{-0.2cm}
   \caption{Cartoon faces of various groups. The female's faces have bright eyes with long eyelashes while the male's do not. Furthermore, the kids' faces have obvious blush and the elderly's faces are wrinkled. The real-faces and the cartoons do not have one-to-one correspondence, as we do not require paired training samples.}
\label{fig:fig1}
\vspace{-0.2cm}
\end{figure}

In this paper, we focus on "face-to-cartoon" conversion. Several previous works \cite{FaceGAN, U-GAT-IT} have studied this problem and achieved impressive results. However, the cartoon style in these works is relatively monotonous. They do not consider the difference among various groups. For example, as can be seen in Figure \ref{fig:fig1}, the female's cartoon faces have big eyes with long eyelashes while the male's may not have eyelashes. The elderly's faces are usually wrinkled but the kids' are not. Therefore for better performance, it is necessary to consider such differences when collecting training samples and designing models. For data collection, real-face images are easy to obtain on the Internet, while cartoon-faces of specific style are hard to collect or design. In addition, the amounts of available cartoon-faces of each group are also unbalanced. Cartoon faces of young women are more common than other groups on the web.

Thus we propose a method for fine-grained face-to-cartoon generation. Our key assumption is, although different groups have their specific appearance, they still share the same cartoon style. As a result we can first train an image translation model for the common group(the female) and then transfer knowledge to other groups with only few samples. We design a multi-branch image translation network for fine-grained face generation. The main branch learns to translate images from the common group and maintain the distribution of the shared feature space while other branches learn specific characteristics for each rare group. In this manner, we can learn a few-shot image translation model effectively. The main contribution of the proposed work can be summarized as follows:

\begin{itemize}
\item We propose a two-stage training strategy and a multi-branch image translation model for few-shot fine-grained cartoon face generation.
\item We collect a cartoon-face dataset for such a fine-grained image translation task, which consists of four groups: young women, young men, the kids and the elderly.
\item We achieve good performance with only few samples for specific groups.
\end{itemize}

\section{Related works}
\subsection{Generative Adversarial Networks}
Generative Adversarial Networks(GAN) \cite{GAN} have raised a lot of attention since proposed. It has achieved impressive results in various fields, such as image generation \cite{DCGAN, WGAN, KarrasALL18}, image completion \cite{IizukaS017, deepfill1, deepfill2}, image translation \cite{pix2pix, cycleGAN,starGAN, starGAN_v2}, etc. In training, a generator aims to generate realistic images to fool a discriminator while the discriminator tries to distinguish the generated images from real images. In this paper, our model uses GAN to learn the face-to-cartoon translation for multi groups, given unpaired training data.

\subsection{Image-to-Image Translation}
Image-to-image translation aims to transform an image from the source domain to the target domain. It involves supervised and unsupervised translation. For supervised settings, paired training samples are available. Pix2pix \cite{pix2pix} applies adversarial loss with L1-loss to train a condition generative network. High-resolution version of the pix2pix have also been proposed by Wang \etal\cite{HR-pix2pix}. For unsupervised settings, no paired data are available anymore. CycleGAN \cite{cycleGAN}proposes a cycle consistency loss to learn two mapping functions at the same time. UNIT\ cite{UNIT} assumes a shared-latent space to tackle unsupervised image translation. Furthermore, MUNIT \cite{MUNIT} decomposes the image into content code that is domain-invariant and a style code that captures domain-specific properties and can thus extend to many-to-many mappings. pRecently, U-GAT-IT \cite{U-GAT-IT} incorporates a new attention module based on Class Activation Map (CAM) \cite{CAM} to guide the model to focus on more important regions distinguishing between source and target domains. A novel Adaptive Layer-Instance Normalization (AdaLIN) is also proposed to help the attention-guided model to flexibly control the amount of  change in shape and texture. Additionally, some methods \cite{FaceGAN, CartoonGAN, CaoLY18} focus on some specific data form like cartoon. In summary, the methods above all require large amount of data for training, no longer supervised or unsupervised. However, in our setting, we do not have efficient data for some groups.

\subsection{Few-shot Image-to-Image Translation}
Some works have also studied the image-to-image translation task in few-shot settings. Benaim and Wolf \cite{OneShotTrans} propose an unsupervised framework to solve the one-shot image translation task. In their settings, the source domain only contains a single image while the target domain have many images during training. They first train a variational auto-encoder for the target domain. Then given the single training source image, they create a variational auto-encoder for source domain by adapting the low-level layers close to the image in order to directly fit the source image. Cohen and Wolf \cite{BiOneShotTrans} further improve at the same setting. They employ one encoder and one decoder for each domain without utilizing weight sharing. The auto-encoder of the single sample domain is trained to match both this single sample and the latent space of target domain. Liu \etal \cite{FUNIT} study the image translation task in a new few-shot setting. They try to map images in a given class to an analogous image in a different class, which is previously unseen during training. Our setting differs from the above as we assume a group with sufficient training data is available and we need to transfer knowledge to other groups which only contain few training samples.

\section{Our Method}
\subsection{Problem Formulation}
Our goal is to train a generator $T$ that learns mappings between two domains for multi-groups. We denote real-face images and cartoon-face images as domain $\mathcal{X}$, $\mathcal{Y}$ separately. It is worth noting that no paired data is required. Meanwhile, we further divide human beings into four specific groups, young women, young men, kids and the elderly. For convenience, we will refer to these groups as \textit{group 0, 1, 2, 3} in the following paragraphs, denoted as $\mathcal{X}=\{\mathcal{X}_i\}$, $\mathcal{Y}=\{\mathcal{Y}_i\}, i=0,1,2,3$. In these groups, \textit{group 0} contains sufficient training samples, while others only have few images. The cartoon faces of all these groups share the same style but differ in details. Thus we first train a mapping network $T_0$ for group $\{\mathcal{X}_0, \mathcal{Y}_0\}$ and transfer to other groups $\{\mathcal{X}_i, \mathcal{Y}_i\}, i=1, 2, 3$.

\begin{figure}[h]
\begin{center}
\includegraphics[width=0.9\linewidth]{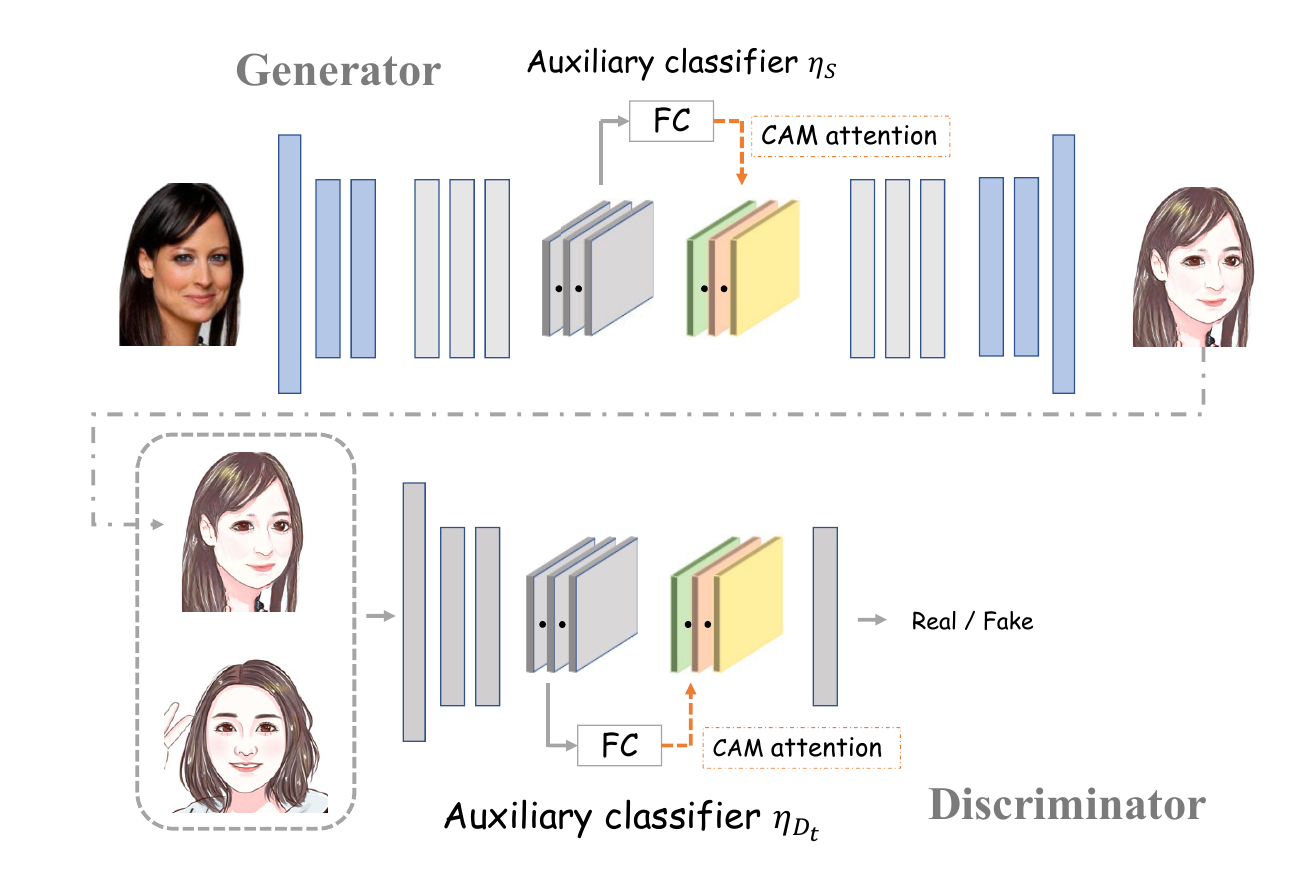}
\end{center}
\vspace{-0.2cm}
   \caption{Basic model architecture.}
\label{fig:pip1}
\vspace{-0.2cm}
\end{figure}

\begin{figure*}[t]
\begin{center}
\includegraphics[width=1.0\linewidth]{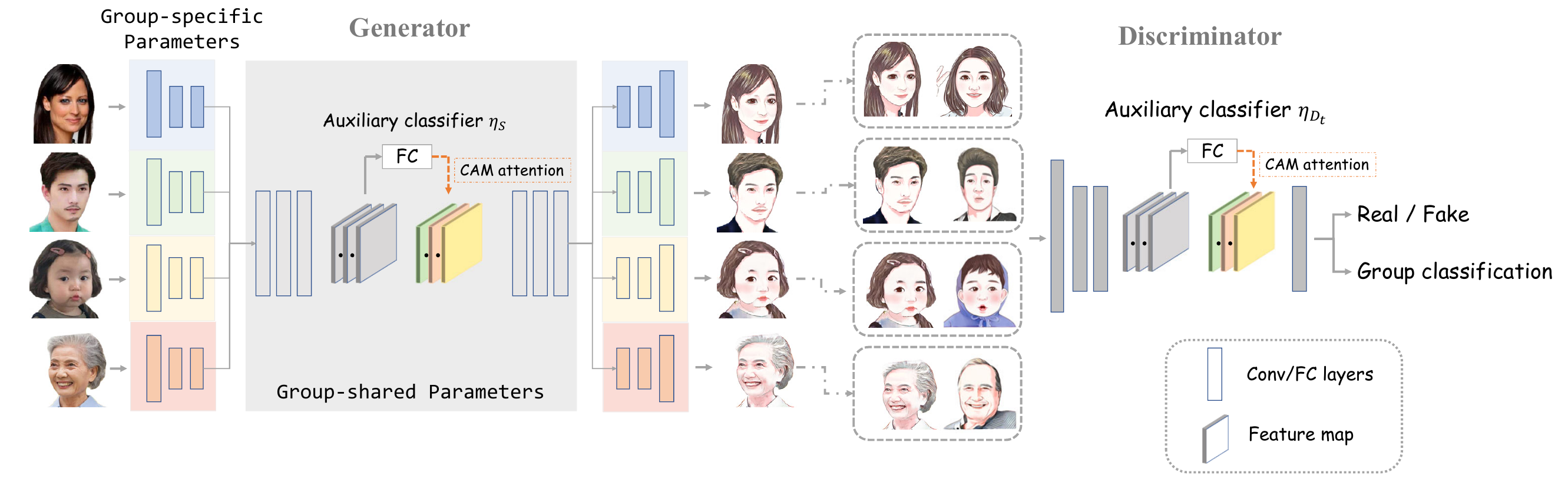}
\end{center}
\vspace{-0.2cm}
   \caption{The whole pipeline. The grey part of the generator represents group-shared parameters. The blue, green, yellow, red ones represent group-specific parameters for young women, young men, kids and the elderly separately. All these parameters are initialized with pre-trained parameters of the basic model.}
\label{fig:pip2}
\vspace{-0.2cm}
\end{figure*}
\subsection{Basic Model}
As mentioned above, we first train an image translation model with $\{\mathcal{X}_0, \mathcal{Y}_0\}$. Some unsupervised methods like CycleGAN \cite{cycleGAN}, UNIT \cite{UNIT} are all suitable. Here we utilize an open source project\footnote{https://github.com/minivision-ai/photo2cartoon\label{web}} for its impressive performance, which is modified from U-GAT-IT\cite{U-GAT-IT}.

The basic model architecture can be seen in Figure \ref{fig:pip1}, it follows an Encoder-Decoder structure, consisting of several down-sampling/up-sampling blocks, hourglass blocks and res-blocks. In generator, a new attention module based on Class Activation Map (CAM) \cite{CAM} is used to guide the model to focus on more important regions by the auxiliary classifier, which helps distinguish between source and target domains. And in discriminator, another auxiliary classifier helps distinguish between real and fake images. In project \ref{web}, a face-ID loss are introduced to further improve the performance in response to faces. More details can be found on the project homepage \ref{web}.

\subsection{Knowledge Transfer for Few-shot Learning} \label{sec:method}
After training an encoder-decoder generator for \textit{group 0}, we then focus on how to transfer knowledge and learn mapping functions for other groups with only few samples. We tackle this problem as a domain adaptation task among various groups.

As discovered in \cite{OneShotTrans}, for few-shot learning, the low layers of encoder and the top layers of decoder can be domain-specific and unshared, while the encoder's top layers and the decoder's bottom layers should be shared. They find that updating the whole network for few-shot samples would quickly lead to overfitting and unstability during training. However, if the shared representation is completely fixed, the lack of adaptation hurts the performance as well. Thus a selective-back propagation mechanism is proposed to solve the one-shot problem. Here we borrow this idea and propose a multi-branch network architecture for our few-shot fine-grained cartoon face generation.

The whole pipeline can be seen in Figure \ref{fig:pip2}. Based on the generator trained for \textit{group 0}, we first add three group-specific branches for other groups each. These new branches are initialized with pre-trained parameters of the basic model.

We denote the group-specific encoder layers as $E^i_{s \rightarrow t}$ and decoder layers as $G^i_{s \rightarrow t}$ for \textit{group $i$, $i=0,1,2,3$}. And the group-shared encoder layers are represented as $E^S_{s \rightarrow t}$, decoder layers as $G^S_{s \rightarrow t}$. For convenience, we denote source domain as $s$ and target domain as $t$ here. They can both be real-face images $\mathcal{X}$ or cartoon-face $\mathcal{Y}$. Then for a image $x$ in \textit{group i}, the mapping function from source domain $s$ to target domain $t$ $T^i_{s \rightarrow t}$ can be denoted as:
\begin{equation*}
T^i_{s \rightarrow t} = G^i_{s \rightarrow t}(G^S_{s \rightarrow t}(E^S_{s \rightarrow t}(E^i_{s \rightarrow t}(x)))), i = 0, 1, 2, 3.
\end{equation*}

For our few-shot multi-group translation task, $E^S_{s \rightarrow t}$ and $G^S_{s \rightarrow t}$ enforce the same structure and style of all the groups. We assume that it is sufficient to maintain the shared feature distribution using \textit{group 0}, as it contains large numbers of training samples. The few samples in other groups are only used to update their specific parameters. That means, the shared parameters $E^S_{s \rightarrow t}$ and $G^S_{s \rightarrow t}$ are detached from the losses on \textit{group 1, 2, 3}, as loss items on few-shot groups may hurt the shared feature space.

Let $x$ as a training sample, then the multi-group encoding/decoding can be denoted as:
\begin{equation*}
T^0_{s \rightarrow t} = G^0_{s \rightarrow t}(G^S_{s \rightarrow t}(E^S_{s \rightarrow t}(E^0_{s \rightarrow t}(x))));
\end{equation*}
\begin{equation*}
T^1_{s \rightarrow t} = G^1_{s \rightarrow t}(\overline{G^S_{s \rightarrow t}}(\overline{E^S_{s \rightarrow t}}(E^1_{s \rightarrow t}(x))));
\end{equation*}
\begin{equation*}
T^2_{s \rightarrow t} = G^2_{s \rightarrow t}(\overline{G^S_{s \rightarrow t}}(\overline{E^S_{s \rightarrow t}}(E^2_{s \rightarrow t}(x))));
\end{equation*}
\begin{equation*}
T^3_{s \rightarrow t} = G^3_{s \rightarrow t}(\overline{G^S_{s \rightarrow t}}(\overline{E^S_{s \rightarrow t}}(E^3_{s \rightarrow t}(x)))).
\end{equation*}
where the bar is used to indicate a detached clone not updated during backpropagation.

\subsection{Loss Function}
The full objective of the model comprises six loss functions, four of which are similar in \cite{U-GAT-IT}. Denote source domain distribution as $X_s$, target domain distribution as $X_t$ and the discriminators for source and target domain as $D_s$, $D_t$, the loss items of p$s \rightarrow t$ can be written as:

\textbf{Adversarial loss} An adversarial loss is employed to match the distribution of the translated images to the target image distribution:
\begin{equation*}
L_{adv}^{s \rightarrow t} = (\mathbb{E}_{x \thicksim X_t}[(D_t(x))^2] + \mathbb{E}_{x \thicksim X_s}[(1-D_t(T^i_{s \rightarrow t}(x)))^2]).
\end{equation*}

\textbf{Cycle loss} A cycle consistency is applied to constraint to the generator as CycleGAN\cite{cycleGAN} to alleviate the mode collapse problem:
\begin{equation*}
L_{cycle}^{s \rightarrow t} = \mathbb{E}_{x \thicksim X_s}[
\lvert x - T^i_{t \rightarrow s}(T^i_{s \rightarrow t}(x)) \rvert _1].
\end{equation*}

\textbf{Identity loss} An identity consistency constraint is used to ensure that the color distributions of input image and output image are similar:
\begin{equation*}
L_{identity}^{s \rightarrow t} = \mathbb{E}_{x \thicksim X_t}[
\lvert x - T^i_{s \rightarrow t}(x) \rvert _1].
\end{equation*}

\textbf{CAM loss} By exploiting the information from the auxiliary classifier $\eta_{s}$ and $\eta_{D_t}$, given an image $x \in \{\mathcal{X}_i\}, i=0,1,2,3$, $T^i_{s \rightarrow t}$ and $D_{tp}$ get to know where they need to improve or what makes the most difference between two domains in the current state:
\begin{equation*}
L_{cam}^{s \rightarrow t} = -(\mathbb{E}_{x \thicksim X_s}[log(\eta_{s}(x))] + \mathbb{E}_{x \thicksim X_t}[log(1 - \eta_{D_t}(x))]),
\end{equation*}
\begin{equation*}
L_{cam}^{D_t} = \mathbb{E}_{x \thicksim X_t}[\eta_{D_t}(x)^2] + \mathbb{E}_{x \thicksim X_s}[(1 - \eta_{D_t}(T^i_{s\rightarrow t}(x)))^2].
\end{equation*}

\textbf{Face ID loss} To enforce the corresponding constraints between real faces and cartoon faces, the cosine distance of features between real-face image and cartoon-face image is used:
\begin{align*}
L_{face} = &\mathbb{E}_{x \thicksim X_s}[1 - \cos(F(x), F(T^i_{s \rightarrow t}(x)))] + \\
&\mathbb{E}_{x \thicksim X_t}[1 - \cos(F(x), F(T^i_{t \rightarrow s}(x)))],
\end{align*}
where $F$ represents a pre-trained face recognition model.

\textbf{Group classification loss} For a given image x in \textit{group i}, we add another auxiliary classifier on top of discriminator and impose the group classification loss when optimizing both discriminator and generator. This objective is composed into two items: a group classification loss of real images used to optimize $D_t$, and a group classification loss of fake images used to optimize $T^i_{s\rightarrow t}$:
\begin{equation*}
L_{cls}^{real} = \mathbb{E}_{x \thicksim \mathcal{Y}_i}[-log D^{cls}_t(i|x)],
\end{equation*}
\begin{equation*}
L_{cls}^{fake} = \mathbb{E}_{x \thicksim \mathcal{X}_i}[-log D^{cls}_t(i|T^i_{s\rightarrow t}(x))],
\end{equation*}
where the term $D^{cls}_t(i|x)$ represents a probability distribution over group labels computed by $D^t$.

\textbf{Full objective} Finally, the full objective to train the encoders, decoders, discriminators, and auxiliary classifiers is:
\begin{align*}
L_D &= L_{adv} + L_{cls}^{real} \\
L_G &= \lambda_1 L_{adv} + \lambda_2 L_{cycle} + \lambda_3 L_{identity} \\
&+ \lambda_4 L_{cam} + \lambda_5 L_{face} + \lambda_6 L_{cls}^{fake}
\end{align*}
where $\lambda_1=1$, $\lambda_2=10$, $\lambda_3=10$, $\lambda_4=1000$, $\lambda_5=1$, $\lambda_6=100$. Here, $L_{adv} = L_{adv}^{s \rightarrow t} + L_{adv}^{t \rightarrow s}$ and the other losses are defined in the similar way($L_{cycle}, L_{identity}, L_{cam}$).

\section{Experiments}
\subsection{Datasets}
In order to accomplish our task, we need two domains of data for cartoon and human faces of various groups. We collect these data from existed datasets and the Internet. In addition, we employ a professional painter to draw some cartoon faces for few-shot groups.

For young women, we collect 1,000 real-face images from CelebA dataset \cite{celebA} and 200 cartoon-face images from this open-source project \ref{web}. For other groups, we collect 75 real-face images for young men, 50 for the kids and 50 for the elderly. We also employ a painter to draw 15 cartoon faces for young men, 10 for the kids and 10 for the elderly.

Follow the data pre-process procedure in project \ref{web}, face alignment and semantic segmentation are used to help reduce the difficulty of optimization.

\subsection{Implementation Details}
The whole project is built based on project \ref{web}. All models are trained using Adam\cite{Adam} with $\beta_1=0.5$ and $\beta_2=0.999$. For data augmentation, images are resized to $286\times286$ and random cropped to $256\times256$. We train all models with a fixed learning rate of 0.0001 and use a weight decay at rate of 0.0001.  The weights are initialized from a zero-centered normal distribution with a standard deviation of 0.02 for the basic model. For few-shot transfer model, the weights are initialized from the basic model. The batch size is set to one for each group. That means, we sample one image from each group once and forward the network together. During backpropagation, selective backpropagation is used as described in section \ref{sec:method}.

\subsection{Ablation Study}
\subsubsection{Compared with several Simple Baselines}
We compare our method with several simple baselines to verify the effectiveness.

\begin{figure}[h]
\begin{center}
\includegraphics[width=0.9\linewidth]{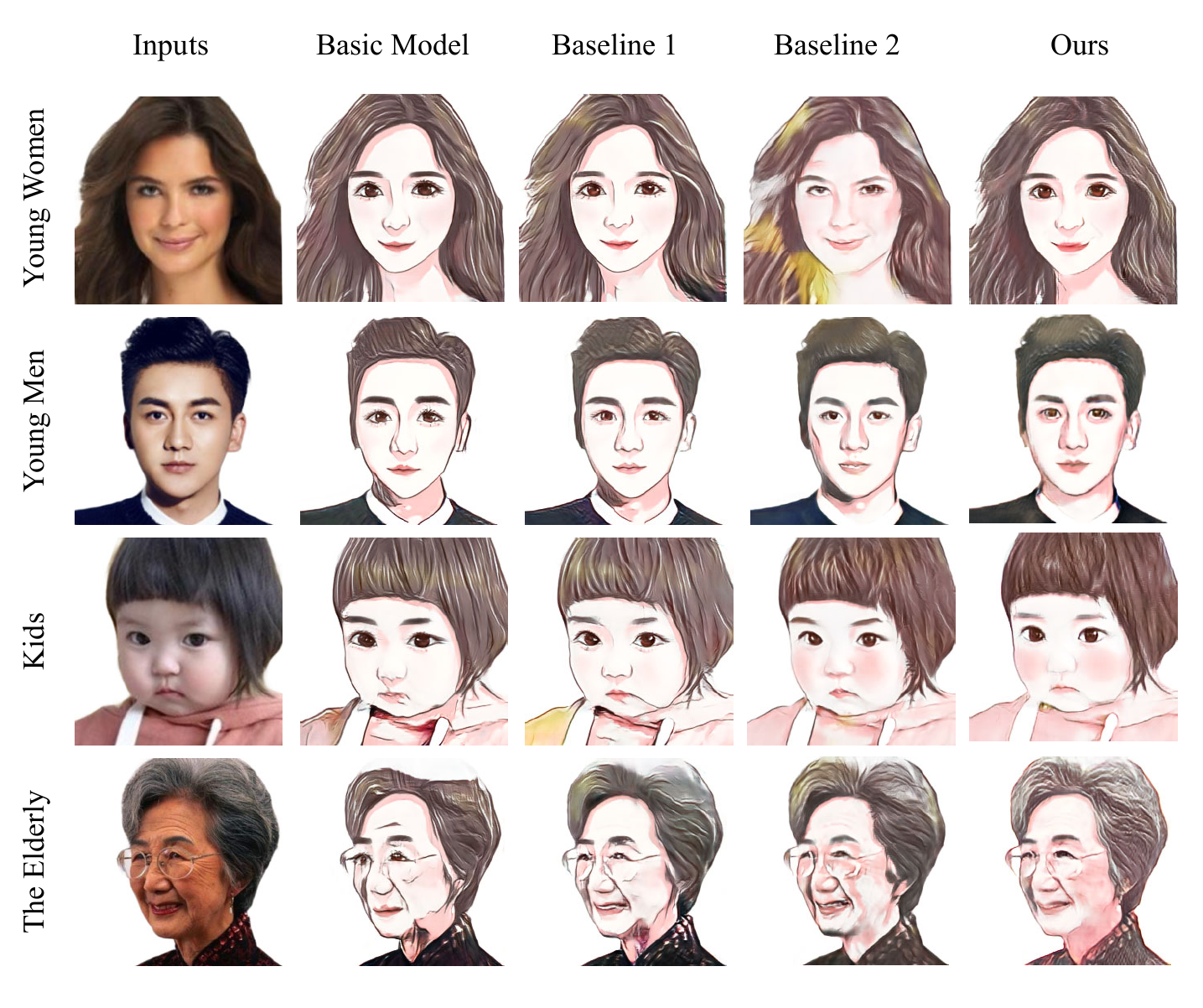}
\end{center}
\vspace{-0.2cm}
   \caption{Compared with several simple baselines.}
\label{fig:cmp3}
\vspace{-0.2cm}
\end{figure}

\textbf{Basic Model.} The basic model is trained on the group of young women. Some results of each group can be seen in the second column of Figure \ref{fig:cmp3}. The generated cartoon face of woman is pretty good. However, when it comes to other groups, some problems will arise. For example, the man's portrait has long eyelashes, which makes it seems feminine. For the old lady, as the basic model never sees any samples with deep wrinkles, the result is obvious strange.

\textbf{Baseline 1.} In project \ref{web}, they train the basic model with about 200 cartoon pictures for each group while we have only few samples for most groups. For comparison, we train a baseline by mixing all the training data, ignoring the differences among groups. Results can be seen in the third column of Figure \ref{fig:cmp3}. Compared with basic model, the results of men and kids seem to be improved a little. However, as samples of young women are in the majority during training, results of other groups still seems feminine.

\textbf{Baseline 2.} In this setting, we directly finetune the basic model on other groups, ignoring group labels. That means we combine the few-shot samples from young men, kids and the elderly for finetuning. Feminization is not observed in the forth column of Figure \ref{fig:cmp3}. Cartoon faces of the man and the kid are greatly improved compared with previous results. Nevertheless, the portrait of the woman is of bad quality, as the model learns to transfer to other groups and forgets the knowledge of the previous group.

\begin{figure}[t]
\begin{center}
\includegraphics[width=0.9\linewidth]{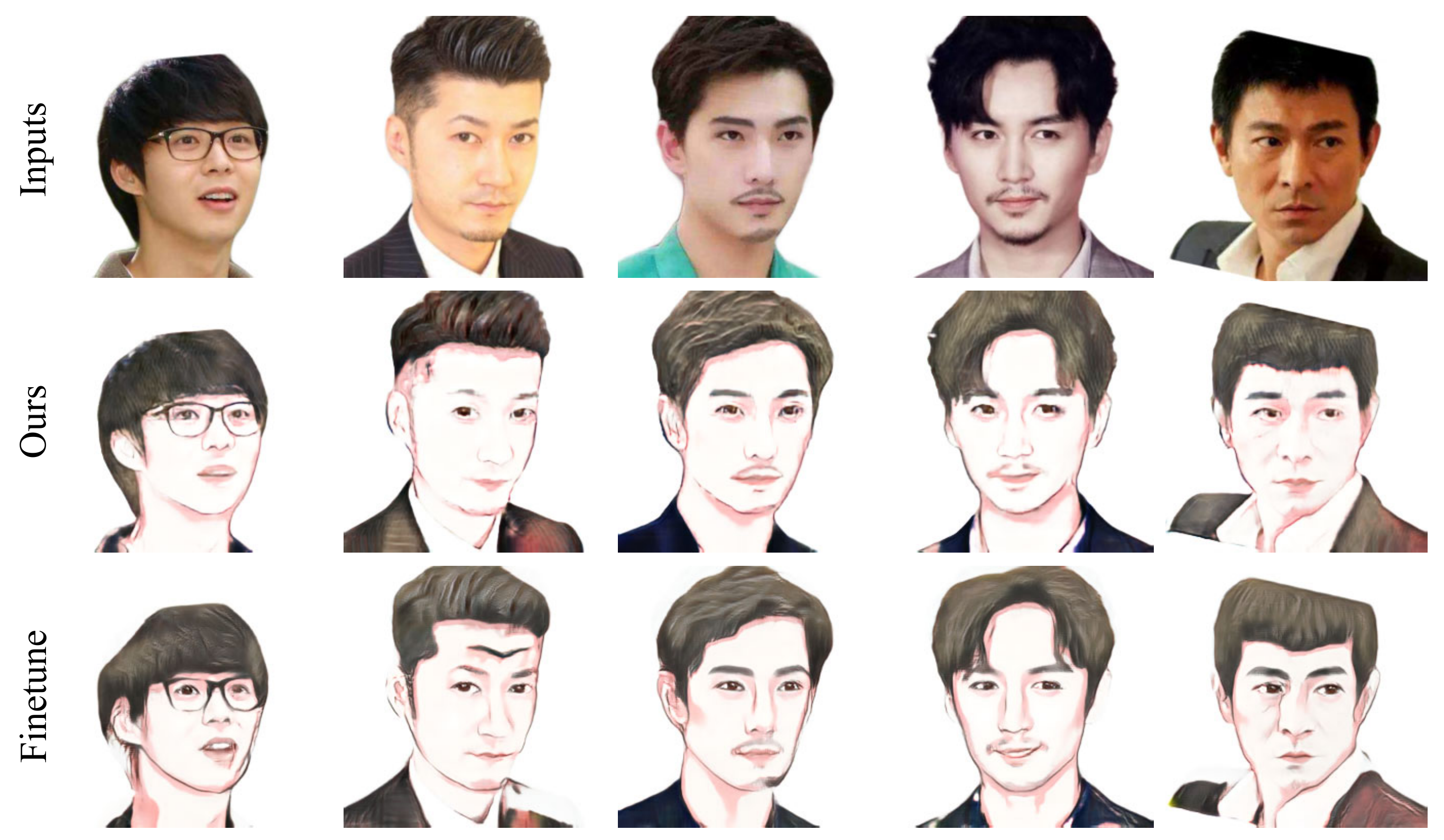}
\end{center}
   \caption{Compared with directly finetune on the group of young men. A better correspondence can be seen in the second row.}
\label{fig:cmp1}
\vspace{-0.5cm}
\end{figure}

\textbf{Baseline 3.} In baseline 2, we mix the few-shot samples of different groups. However, a better manner may be that we finetune the basic model on each group separately. As we have three few-shot groups here, for simplification, we only finetune on the group of young men. The results of finetuned model (the third row) in Fiture \ref{fig:cmp1} are a bit unstable. We believe that it is due to the few training data will lead to overfit on specific samples. Our results(the second row) can be seen better corresponded with the input images,
especially from the mouth in the forth and fifth column.

\textbf{Conclusion.} From the results of the above, we can see that it is necessary to design specific representations or models for each group. Meanwhile, the roles of the common group and the few-shot groups also need to be carefully considered. From the fifth column of Figure \ref{fig:cmp3}, we can see our full model successfully capture the group-specific characteristics.

\subsubsection{Compared with Training Without Selective Backpropagation}
As discussed in section \ref{sec:method}, we only update group-specific parameters for few-shot groups but update the whole network for the common group. In this experiment we remove selective backpropagation during training, that means we do not distinguish samples from few-shot groups or the common group and update the group-shared parameters for few-shot samples as well.

\begin{figure}[h]
\begin{center}
\includegraphics[width=0.9\linewidth]{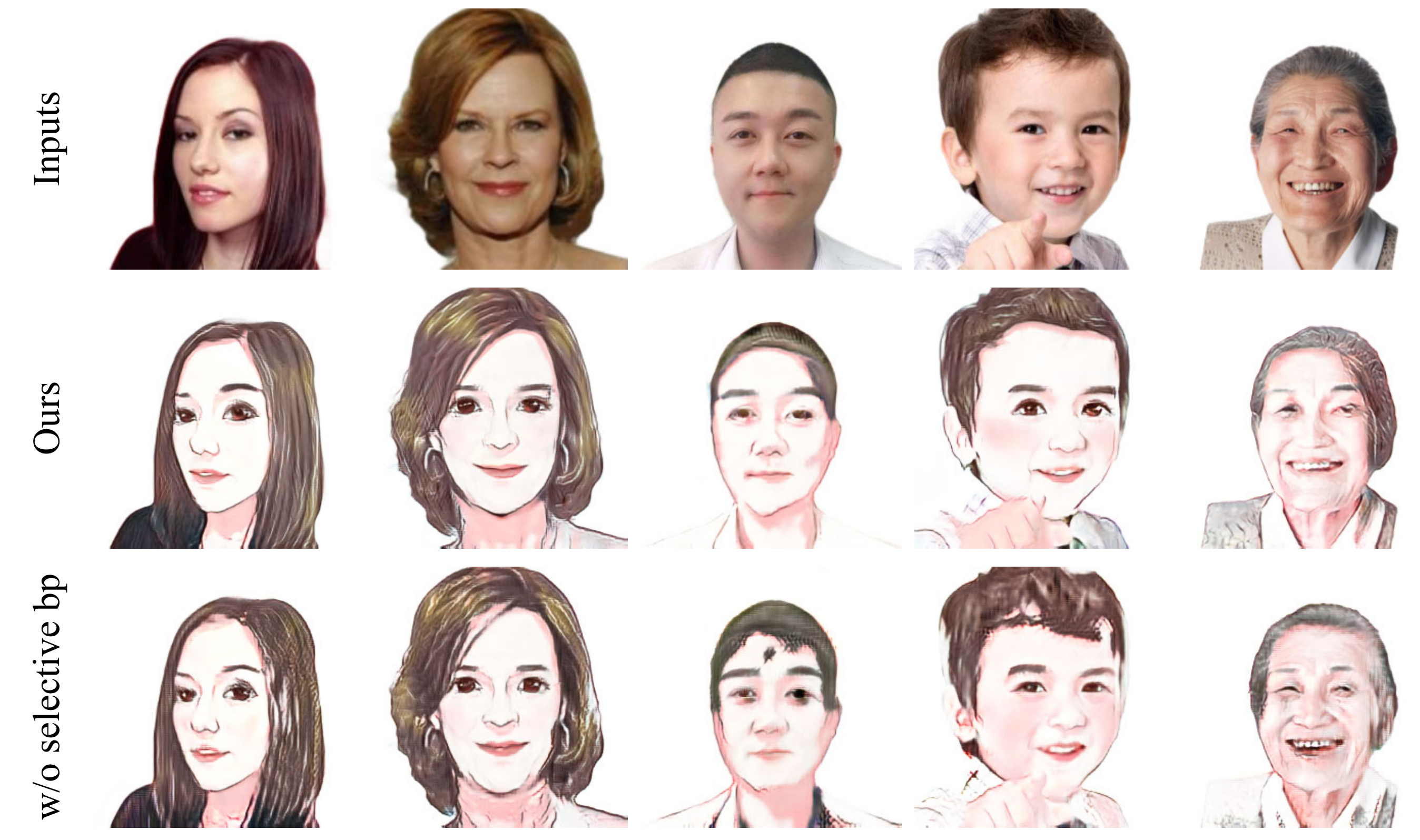}
\end{center}
\vspace{-0.2cm}
   \caption{Compared with training without selective backpropagation. Obvious fake patterns can be seen in the third row.}
\label{fig:cmp2}
\vspace{-0.2cm}
\end{figure}

In Figure \ref{fig:cmp2}, we can see obvious fake patterns in the third row. Some wrinkles are observed on the neck of the two women. We believe this is in that during training, the common group and the few-shot groups have equal contribution to the shared feature space, which may hurt the distribution. Similarly, for results of men, kids and the elderly, the results of our method are much better.

\subsection{Analysis of CAM Attention}
\begin{figure}[h]
\begin{center}
\includegraphics[width=0.9\linewidth]{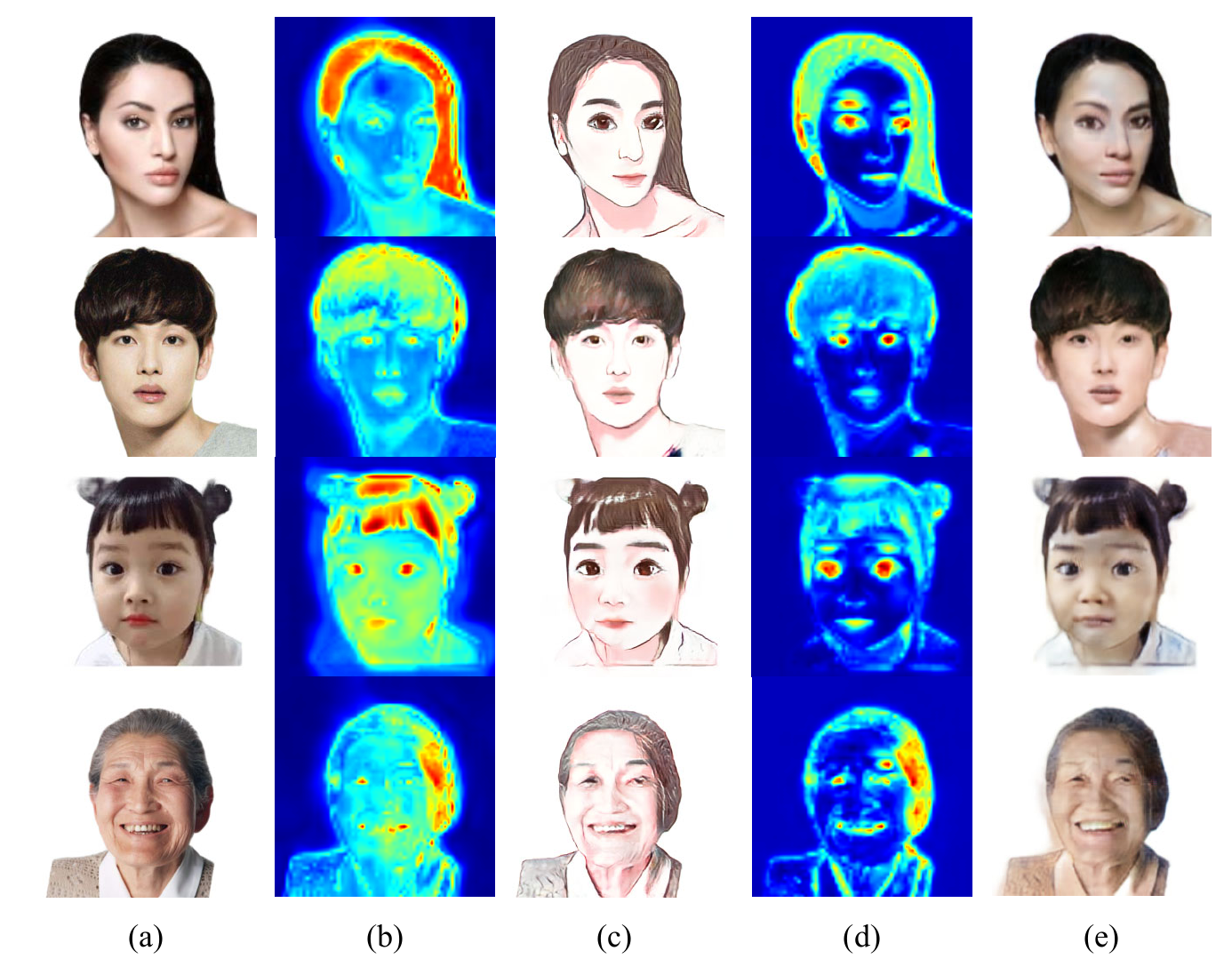}
\end{center}
\vspace{-0.2cm}
   \caption{Visualization of the attention maps: (a) Real-face images, (b) Attention map of generator from real-face to cartoon-face, (c) Generated cartoon-face images, (d) Attention map of generator from generated cartoon to real faces; (e) Reconstruction of real faces.}
\label{fig:att}
\vspace{-0.2cm}
\end{figure}

We visualize some attention maps learned by the generator in Figure \ref{fig:att}. As our model is unsupervised, there are two generators to maintain a cycle-consist constraint. In column (b) of Figure \ref{fig:att}, we can find that during face-to-cartoon generation, the attention maps tend to focus on the whole face, especially the hair and eyes. This is consistent with our intuition, as the color distribution and details of hair and eyes are significant and distinguishable between the two domains.

However, when it comes to translation from cartoon to real-face images, the response areas in the attention maps are much smaller but focus more around eyes and hair, as seen in column (d). We believe this is because that the variance of color distribution of cartoon faces is less detailed than that of real faces. When real faces are translated to cartoons, some color details are lost to some extent. Thus when translated back to real-face images, the model focuses more on some explicit regions, such as eyes and hair.

\subsection{Cartoon Face Generation}
We show some face-to-cartoon generation results in Figure \ref{fig:res1}, Figure \ref{fig:res2}, Figure \ref{fig:res3}, Figure \ref{fig:res4}. As can be seen, our model captures specific characteristics for each group well, such as long eyelashes in Figure \ref{fig:res1}, obvious blush in Figure \ref{fig:res3} and wrinkles in Figure \ref{fig:res4}. These differences make the cartoon of various groups more fine-grained.

\subsection{Failure Cases}
\begin{figure}[h]
\begin{center}
\includegraphics[width=0.9\linewidth]{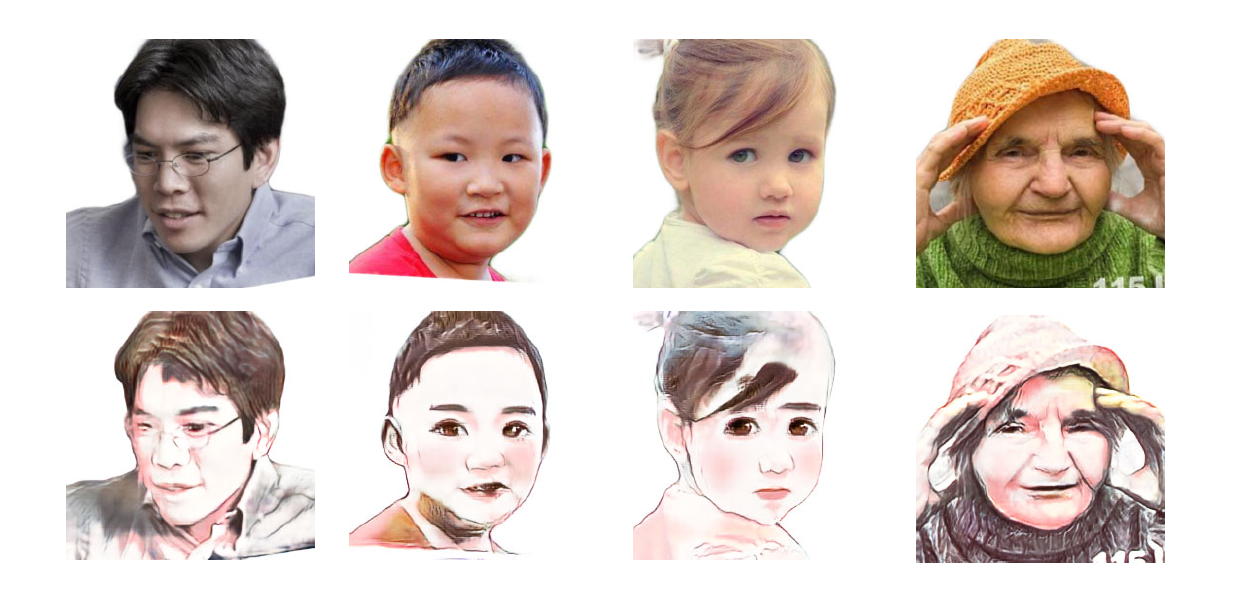}
\end{center}
\vspace{-0.2cm}
   \caption{Some failure cases.}
\label{fig:fail}
\vspace{-0.2cm}
\end{figure}

\begin{figure*}[p]
\begin{center}
\includegraphics[width=1.0\linewidth]{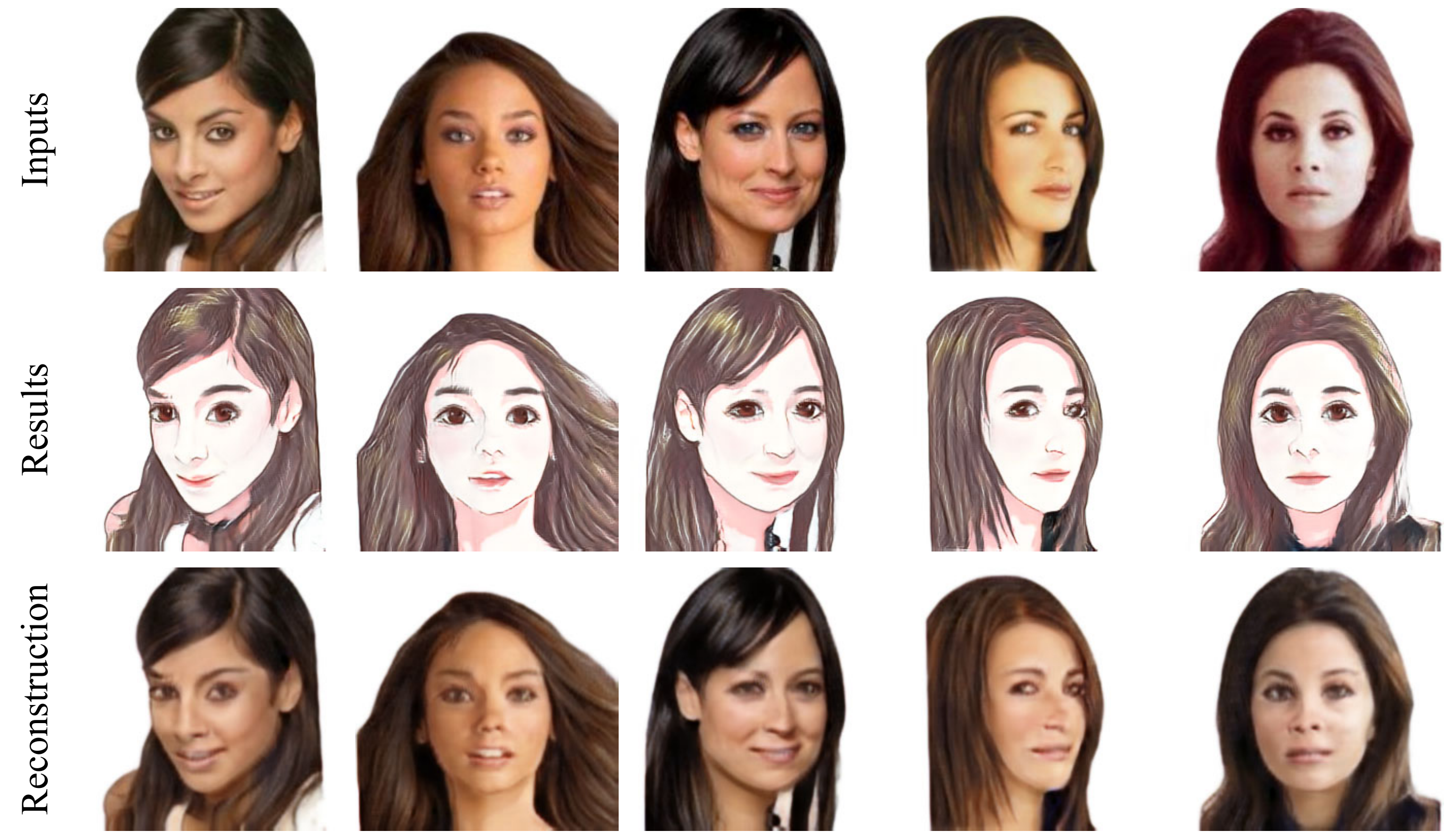}
\end{center}
   \caption{Results of young women.}
\label{fig:res1}
\end{figure*}

\begin{figure*}[p]
\begin{center}
\includegraphics[width=1.0\linewidth]{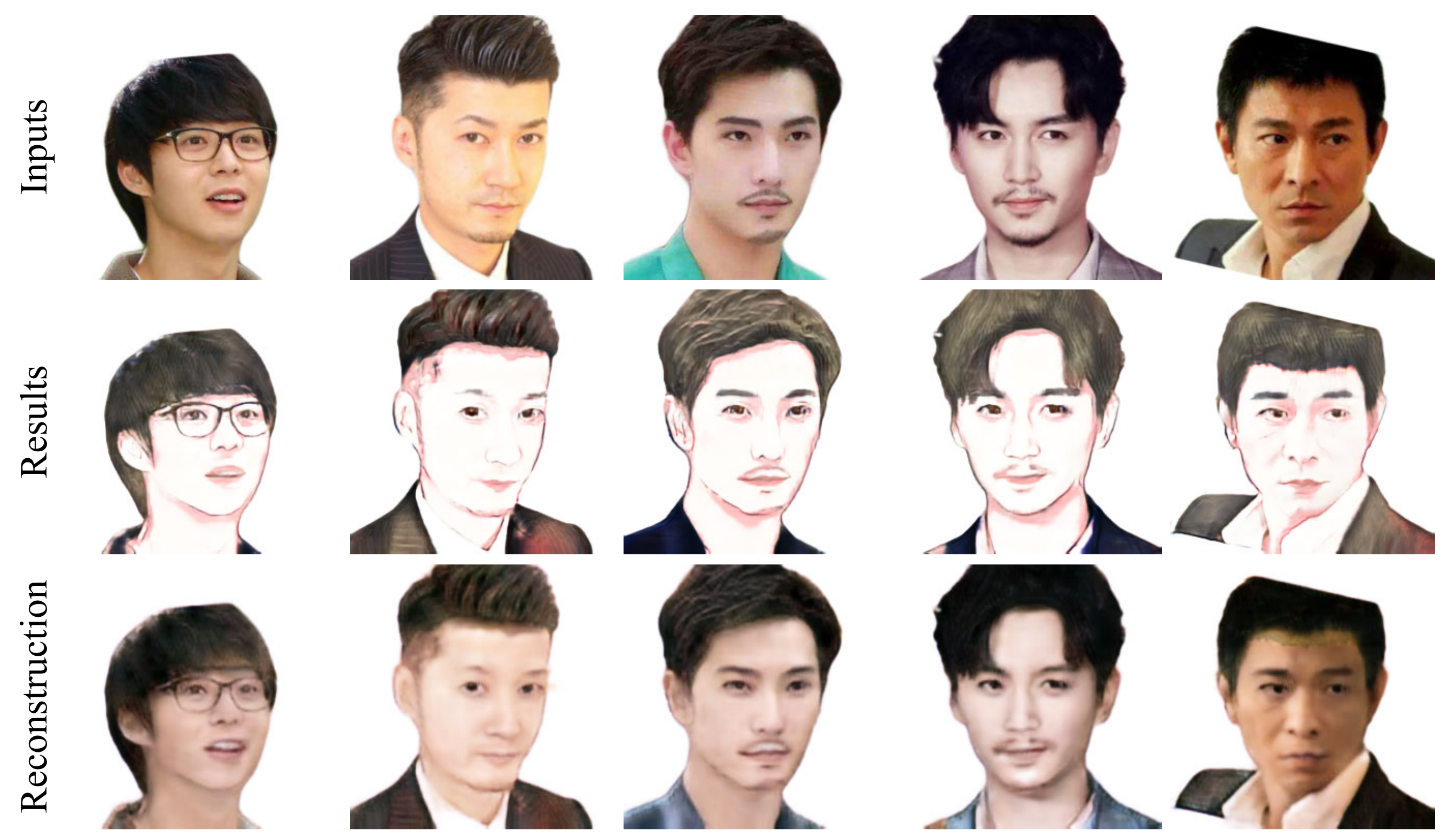}
\end{center}
   \caption{Results of young men.}
\label{fig:res2}
\end{figure*}

\begin{figure*}[p]
\begin{center}
\includegraphics[width=1.0\linewidth]{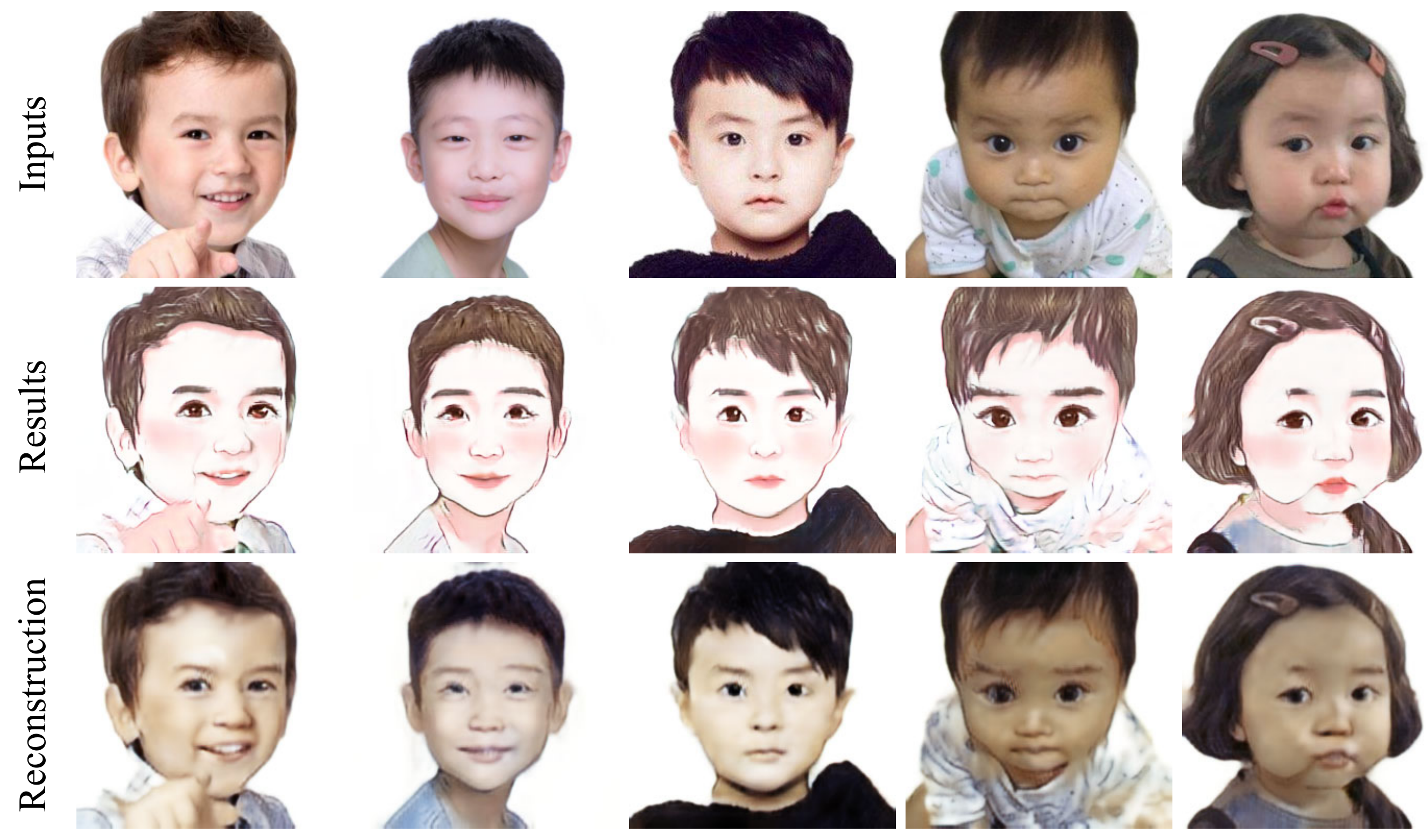}
\end{center}
   \caption{Results of kids.}
\label{fig:res3}
\end{figure*}

\begin{figure*}[p]
\begin{center}
\includegraphics[width=1.0\linewidth]{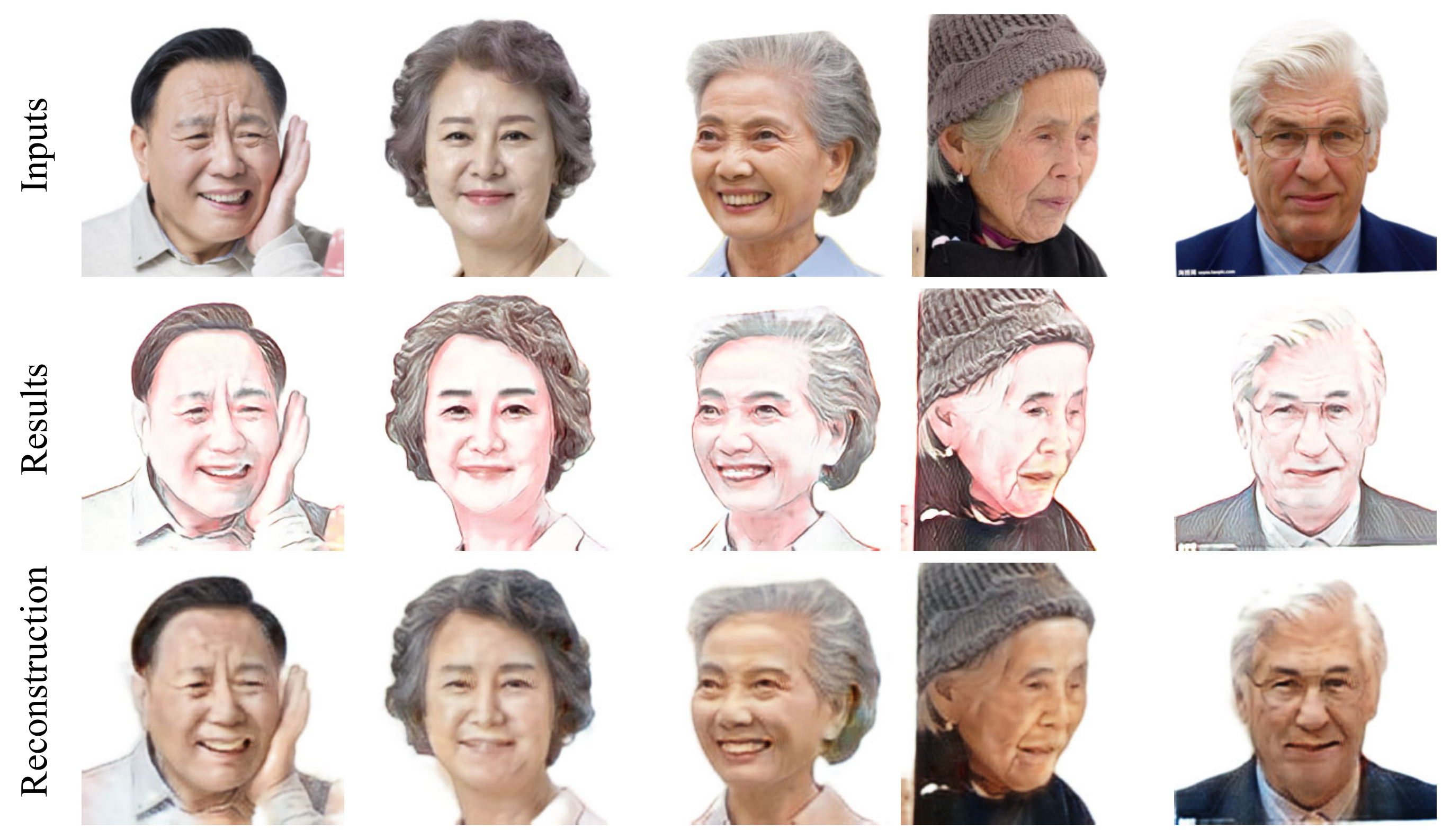}
\end{center}
   \caption{Results of the elderly.}
\label{fig:res4}
\end{figure*}

In Figure \ref{fig:fail} we show some failure cases. Sometimes there are still some fake patterns due to the shadow or complicated contents on faces, such as teeth. These failure cases are caused by the lack of diversity during training. We believe it will be improved if we carefully collect and select training samples.

In addition, as the hair color of cartoon is almost fixed during training, when real-face image has a different hair color with other training samples, the generated results may be unstable, like the third column in Figure \ref{fig:fail} shows.

\section{Conclusion and future work}
In this work, we study the problem of fine-grained cartoon face generation in the few-shot setting. We propose a two-stage training procedure for this problem and achieve good results. However, the results still have some problems. Although our method can generate fine-grained cartoon faces, the details of some key components seems a little blurry when looking closer. Thus in the future, we will consider to improve these details by introducing a novel attention block to capture the differences among groups explicitly.

\section{Acknowledgements}
This project is supported by ByteDance AI Lab. And thanks Wenqi Li and Yuning Jiang for their suggestions on designing cartoon faces of specific groups.

\clearpage
{\small
\bibliographystyle{ieee_fullname}
\bibliography{egbib}
}
\end{document}